\documentclass[runningheads]{llncs}

\usepackage{eccv}

\usepackage{eccvabbrv}

\usepackage{graphicx}

\usepackage{url}
\usepackage{caption}
\usepackage[table]{xcolor}
\usepackage{makecell} 
\usepackage{pifont}
\usepackage{amsmath}
\usepackage{algpseudocode}
\usepackage{bbm}
\usepackage{mathtools}
\usepackage{wrapfig}    
\usepackage{booktabs}   
\usepackage{multirow}   
\usepackage{array}      
\usepackage{amssymb}    
\usepackage{capt-of} 
\usepackage{xcolor,tabularx}
\usepackage{algorithm}
\usepackage{amsmath} 

\usepackage[accsupp]{axessibility}  
\usepackage{pifont}  

\usepackage{hyperref}

\usepackage{orcidlink}

\newcommand{\myparagraph}[1]{\noindent \textbf{#1}}

\begin{document}

\title{PixelPilot: Scalable Vision-Language-Action Models for End-to-End Autonomous Driving} 

\titlerunning{PixelPilot}

\author{Pin Tang\inst{1} 
\and Guoqing Wang\inst{1} 
\and Xiangxuan Ren\inst{1} 
\and Zhongdao Wang\inst{2} 
\and \\Guodongfang Zhao\inst{2} 
\and Bailan Feng\inst{2} 
\and Chao Ma\inst{1,}\thanks{Corresponding author.}
}

\authorrunning{P.~Tang et al.}

\institute{$^1$ MoE Key Lab of Artificial Intelligence, Institute of AI, \\
Shanghai Jiao Tong University, Shanghai, China \\$^2$ Central Research Institute, Huawei, Beijing, China
{\tt\small \{guoqing.wang,pin.tang,bunny\_renxiangxuan,chaoma\}@sjtu.edu.cn} \\
{\tt\small \{wangzhongdao,zhaoguodongfang,fengbailan\}@huawei.com}\\
{\small Project page: \url{https://pixelpilotvla.github.io/}}
}

\maketitle

\begin{abstract}
Vision-Language-Action Models (VLAs), which leverage the advanced reasoning capabilities of Vision-Language Models (VLMs), show promising generalization in complex autonomous driving scenarios. 
Existing VLAs typically predict and optimize 3D trajectories from 2D images. While intuitive, this 2D-to-3D prediction is inherently entangled with camera parameters, leading to limited data scalability across heterogeneous driving datasets. Moreover, directly optimizing in 3D space induces severe convergence to trivial solutions, where VLAs rely on ego-status rather than visual scene understanding. To address these issues, we propose PixelPilot, a novel VLA featuring a decoupled planning and lifting paradigm.
In the planning phase, PixelPilot reformulates scene understanding and trajectory prediction as sensor-agnostic 2D-to-2D tasks in the image plane, thereby facilitating scalable training across diverse datasets. The planned 2D trajectories are then deterministically lifted to 3D only during inference, ensuring the full exploitation of visual cues and generalization across different vehicles. To realize this paradigm, we propose a knowledge-instilled policy learning strategy that applies dense, intermediate rewards via Group Relative Policy Optimization (GRPO) to enforce a rigorous causal chain from visual perception to spatial planning. Extensive experiments demonstrate that PixelPilot achieves state-of-the-art performance in both open-loop and closed-loop settings, validating its superior scalability and visual reasoning capabilities.

\end{abstract}

\section{Introduction}
\label{sec:intro}

\begin{figure}[t]
    \centering
    \includegraphics[width=1.0\linewidth]{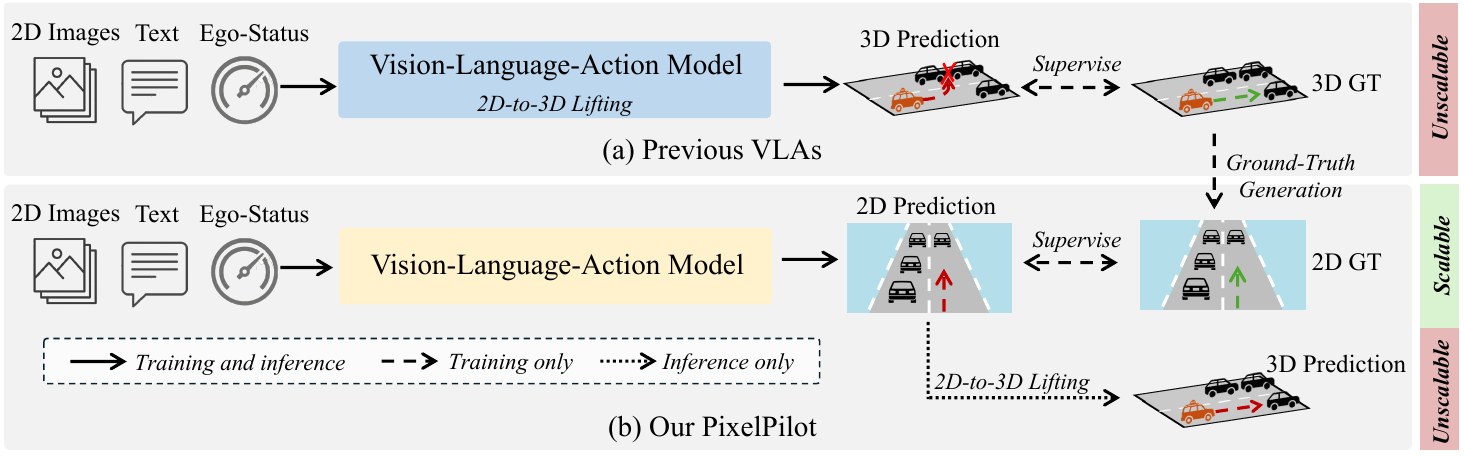}
    \caption{(a) Previous VLAs directly predict 3D trajectories from 2D images which involves learning sensor-specific 2D-to-3D mapping, leading to limited data scalability. (b) Our PixelPilot features a decoupled planning and lifting paradigm. It plans and optimizes 2D trajectories within the image plane, facilitating data scaling and visual reasoning, while unscalable 2D-to-3D trajectory lifting occurs only during inference.
    }
    \label{fig:intro}
\end{figure}

The emergence of Vision-Language Models (VLMs)~\cite{achiam2023gpt4, Qwen2.5VL} has catalyzed a paradigm shift in end-to-end autonomous driving. Unlike conventional methods~\cite{uniad, vad, chen2024vadv2}, which are trained from scratch exclusively on driving datasets, Vision-Language-Action Models (VLAs)~\cite{alphadrive,chi2025impromptu,zheng2025driveagent} leverage VLMs to integrate high-level linguistic reasoning with visual understanding, thereby improving generalization in rare and long-tail scenarios. The primary objective of these driving VLAs is to navigate physical environments by generating safe and drivable 3D trajectories. However, adapting VLMs, which are pretrained on 2D image and text corpora, to the precise 3D metric demands of autonomous driving remains a formidable challenge. To bridge this dimensional gap, existing VLAs fine-tune 2D-pretrained VLMs to directly output 3D world coordinates by incorporating auxiliary 3D modules and instructional data. While intuitive, we observe that this 2D-to-3D prediction paradigm encounters two critical problems.

First, direct 3D prediction significantly restricts the data scalability of driving VLAs. In standard 2D visual tasks (e.g., object detection~\cite{deformabledetr}), scaling is highly effective due to the sensor-agnostic nature of 2D-to-2D prediction. Conversely, predicting precise 3D trajectories directly from 2D visual inputs requires the VLA to either explicitly encode~\cite{wang2025omnidrive,fu2025orion,opendrivevla,zheng2025driveagent} or implicitly learn~\cite{autovla,drivevlm,yuan2025autodrive,zeng2025futuresightdrive,chi2025impromptu} sensor-specific 2D-to-3D mappings within its network weights, as illustrated in Fig.~\ref{fig:intro} (a). Because these mappings are inherently tied to specific camera intrinsics and extrinsics, the learned VLA becomes deeply entangled with the sensor configurations of its training data. Consequently, aggregating massive, heterogeneous driving datasets (e.g., nuScenes and Waymo) introduces severe spatial ambiguity: identical 2D pixel coordinates correspond to entirely disparate 3D physical locations under different camera setups. 

Second, this sensor-specific 3D optimization severely compromises the visual reasoning capabilities of VLAs. Confronted with complex and dataset-specific 2D-to-3D mappings that are challenging to optimize, existing VLAs are highly susceptible to spurious correlations and often converge to trivial solutions~\cite{ego-mlp}. Rather than learning robust 3D scene structures and planning trajectories based on visual inputs, they predominantly rely on ego-status (e.g., speed and acceleration) to directly extrapolate the 3D trajectory, as revealed by empirical evidence that removing ego-status from state-of-the-art VLAs~\cite{wang2025omnidrive,chi2025impromptu} degrades performance far more severely than masking the input images (1.98 vs 0.36 L2 error). 

To address these challenges, we introduce PixelPilot, a novel VLA featuring a \textit{decoupled planning and lifting paradigm}, as illustrated in Fig.~\ref{fig:intro} (b). 
During planning, PixelPilot formulates comprehensive scene understanding and trajectory prediction from input images as 2D-to-2D tasks, which are optimized entirely in the image space in a sensor-agnostic manner. 
Therefore, it can aggregate heterogeneous driving datasets with available ego-motion and camera parameters, facilitating scalable training for robust driving capabilities.
During the lifting phase, these predicted 2D pixel trajectories are deterministically projected into the 3D world coordinate system using the specific vehicle's camera parameters. By decoupling the 2D-to-3D lifting process from the learning objective, absolute ego-speed cannot be trivially mapped to pixel displacement without contextual depth, ensuring that the VLA actively relies on visual understanding and spatial reasoning. Additionally, as the sensor-specific 2D-to-3D lifting occurs only during inference, the learned PixelPilot can generalize to different calibrated vehicles simply by updating the sensor-specific parameters. 
This decoupled design conceptually aligns with human driving: skillful human drivers plan paths based on their 2D visual perspective, regardless of the vehicle type (sensor-agnostic planning), and subsequently translate this plan into physical vehicle control based on their familiarity with the vehicle's dimensions (sensor-specific lifting).

To ensure coherent multi-step reasoning within this 2D pipeline, we propose a knowledge-instilled policy learning strategy. Following ego-centric consistency preprocessing for multi-view input, we first conduct Supervised Fine-Tuning (SFT) to learn foundational driving knowledge. Subsequently, we employ Group Relative Policy Optimization (GRPO) for Reinforcement Learning (RL). Crucially, unlike previous methods~\cite{autovla} that only calculate sparse rewards on final predictions, regardless of the generated Chain-of-Thought (CoT), we assign dense and intermediate rewards to verifiable outputs from perception to planning while leaving free-form reasoning flexible. This structure encourages the final spatial planning to be tied to accurate visual scene understanding.

In summary, our contributions are as follows:
\begin{itemize}
\item We propose a decoupled planning and lifting paradigm that shifts VLA optimization entirely to the image space. It facilitates data scaling across heterogeneous driving datasets with available ego-motion and camera parameters, and prioritizes visual reasoning over trivial ego-status.

\item We introduce a knowledge-instilled policy learning strategy. By employing dense intermediate rewards on verifiable outputs, we effectively mitigate the long-horizon credit assignment problem and encourage a rigorous causal chain from visual perception to final spatial planning.

\item Extensive experiments demonstrate that PixelPilot achieves state-of-the-art performance in both open-loop and closed-loop settings, validating the superior scalability and visual reasoning capabilities of our decoupled paradigm.

\end{itemize}

\section{Related Work}
\myparagraph{Vision-Language Models.} The remarkable reasoning capabilities of Large Language Models (LLMs)~\cite{gpt1, gpt3, touvron2023llama,deepseekr1} have propelled the evolution of Vision-Language Models (VLMs)~\cite{clip,minigpt-4,chu2023mobilevlm}, which fuse visual and textual modalities to construct comprehensive and unified representations. Early milestones such as CLIP~\cite{clip} demonstrated the efficacy of aligning image and text features from independent encoders, facilitating robust zero-shot generalization across image-text pairs. Subsequent architectures, notably LLaVA~\cite{llava}, advanced this paradigm by introducing visual instruction tuning via dedicated vision-language projectors, thereby endowing models with sophisticated multi-modal instruction-following capabilities. More recently, state-of-the-art models like the Qwen-VL series~\cite{Qwen2.5VL} and InternVL~\cite{chen2024internvl,zhu2025internvl3} have adopted more powerful multi-modal architectures. These frameworks achieve unprecedented visual understanding and grounding, unlocking complex open-world applications ranging from object localization and segmentation~\cite{lai2024lisa,lantext4seg,ren2025grounding} to end-to-end autonomous driving~\cite{autovla,wang2025omnidrive,fu2025orion,cai2024driving,fu2024drive,li2024driving,ma2024dolphins,sima2024drivelm,marcu2024lingoqa}.
However, 2D-pretrained VLMs inherently struggle with 3D spatial understanding, particularly when tasked with predicting precise 3D metric trajectories. Moreover, their visual grounding capabilities are expressed through image-space entities like bounding boxes. Preserving this native representation during policy learning is therefore important for effectively exploiting VLMs across driving data collected with diverse sensors.

\myparagraph{Autonomous Driving.} 
Autonomous driving has transitioned from traditional modular pipelines, i.e., perception~\cite{li2022bevformer,huang2021bevdet,bevfusion,wang2023openoccupancy,wang2022detr3d}, motion prediction~\cite{wang2024driving,ettinger2021large}, and planning~\cite{ego-mlp}, to end-to-end learning-based paradigms~\cite{chen2024end,hu2022stp3,uniad,vad,sparsedrive,bev-planner,chen2024vadv2,insightdrivesong2025,han2025dme,bridgead}. 
UniAD~\cite{uniad} pioneered the integration of all sub-tasks into a cascaded framework, yielding substantial improvements over modular baselines. 
Subsequent works~\cite{vad, chen2024vadv2,bridgead} adopt Bird’s-Eye View representations and generate planning trajectories through multi-stage interaction modeling. 
Recently, researchers have increasingly leveraged the broad knowledge and visual reasoning of VLMs to enhance generalization in complex driving scenarios. To adapt the 2D-pretrained VLMs for 3D trajectory prediction, current methods typically incorporate explicit 3D visual encoders or curate 3D driving instructional data, optimizing VLMs in 3D space.
For instance, OmniDrive~\cite{wang2025omnidrive}, Orion~\cite{fu2025orion}, and OpenDriveVLA~\cite{opendrivevla} replace the standard ViT~\cite{dosovitskiy2020image} with a 3D Q-Former~\cite{li2023blip2}, explicitly utilizing camera parameters for 2D-to-3D feature projection.
Conversely, other approaches~\cite{drivemm,drivegpt4,drivemoe,drivevlm,drivemlm,alphadrive,yuan2025autodrive,lmdrive,qian2025agentthink,zheng2025driveagent,zeng2025futuresightdrive,autovla,wang2024drivecot} construct 3D question-answering pairs, fine-tuning VLMs to predict 3D trajectories directly from 2D visual inputs. 
Crucially, in both paradigms, the 2D-to-3D mapping remains strictly entangled with the specific camera parameters of the training datasets. This sensor dependence inherently restricts data scalability across heterogeneous driving corpora and induces convergence to trivial solutions. 

To circumvent these limitations, we propose a \textit{decoupled planning and lifting paradigm}. By performing semantic reasoning and motion planning as sensor-agnostic 2D-to-2D prediction tasks, our approach directly optimizes in the 2D image plane, facilitating scalable training and visual reasoning.

\section{Method}

\subsection{Decoupled Planning and Lifting Paradigm}
\label{sec:paradigm}

\begin{figure}[t]
    \centering
    \includegraphics[width=0.95\linewidth]{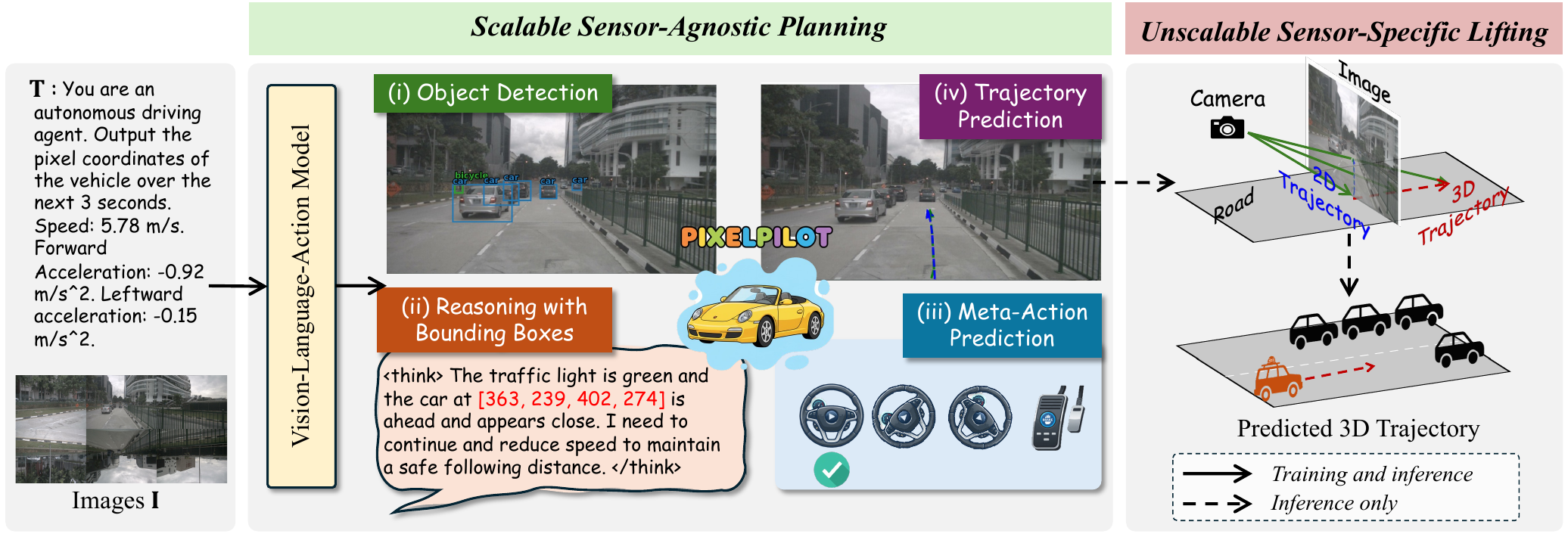}
    \caption{Decoupled planning and lifting paradigm. The sensor-agnostic planning phase is scalable and introduces reasoning with bounding boxes to explore visual cues, while the sensor-specific lifting is unscalable and is performed only at inference.
    }
    \label{fig:paradigm}
\end{figure}

Previous Vision-Language-Action Models (VLAs) are typically fine-tuned to predict 3D trajectories directly. While physically intuitive, this significantly restricts data scaling across heterogeneous driving datasets and degrades visual reasoning, leading to an over-reliance on trivial ego-status. 
To resolve these problems, we introduce PixelPilot, which features a \textit{decoupled planning and lifting} paradigm. As shown in Fig.~\ref{fig:paradigm}, it decouples the pipeline into two phases: 
(1) \textit{planning}, where high-level reasoning and trajectory planning are predicted and optimized entirely in the image space for data scaling and visual reasoning; and (2) \textit{lifting}, where the planned 2D trajectory is deterministically projected into the 3D world for vehicle control.

\begin{figure}[t]
\centering
\captionsetup{type=figure}
\includegraphics[width=0.8\linewidth]{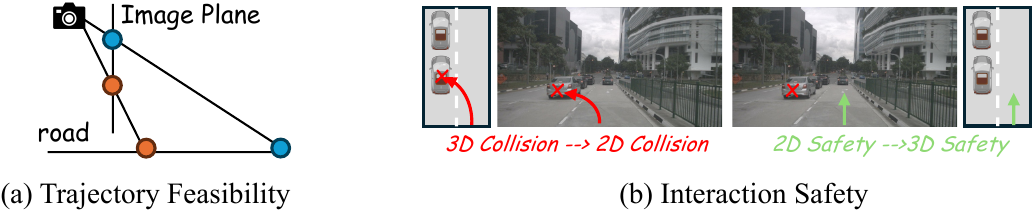}
\caption{Justification for decoupled planning and lifting paradigm. (a) The bijective trajectory pairs demonstrate trajectory feasibility under a local-plane assumption. (b) Planning in image space provides a conservative cue for avoiding visible collision risks.}
\label{fig:justification}
\end{figure}

\myparagraph{Theoretical Justification.} \label{sec:justification}
The core premise of PixelPilot is that shifting trajectory prediction and optimization from 3D space back to 2D images improves data scaling and visual reasoning. 
Before detailing this paradigm, we mathematically justify the rationale of driving in 2D from the following two aspects:

\noindent\textbf{\textit{(1) Trajectory Feasibility.}} For short-term trajectory prediction (e.g., 3 seconds into the future), the road surface immediately ahead, including global edge cases like slopes, can often be approximated as a {local plane}~\cite{lim2021patchwork,lee2022patchwork++} relative to the ego-car as it drives in an ego-centric manner. Our empirical evidence from nuScenes~\cite{caesar2020nuscenes} indicates an average 3-second road height variation of merely 0.16 m, a negligible deviation compared to the 7.6 m depth prediction error typical of expert depth estimators like DepthAnything~\cite{yang2024depthanything}. Consequently, under this local-plane approximation, the 3D road plane and its 2D image projection are related by a strict bijective projective transformation, as visualized in Fig.~\ref{fig:justification} (a). This bijective mapping provides a conditional geometric rationale: every physical coordinate on the 3D drivable road plane maps to a unique, determinable point on its 2D image projection, and vice versa. Therefore, a continuous and smooth trajectory planned on the 2D image corresponds to a physically continuous and smooth trajectory in the 3D metric space under the local-plane assumption.

\noindent\textbf{\textit{(2) Interaction Safety.}} The safety motivation of our 2D planning approach is justified through an analogy to Configuration Space (C-Space) planning~\cite{lozano1983spatial,chen2015deepdriving} in classical robotics. By treating the 2D image plane as a simplified C-space, the 2D bounding boxes of dynamic agents function as Image-Space Obstacles (I-Space Obstacles). Crucially, these I-Space Obstacles represent a conservative superset of visible 3D collision risks. Under accurate visible-object perception and the local-plane assumption, any physical 3D collision implies a 2D projection overlap, so avoiding I-Space Obstacles provides a conservative safety cue, as demonstrated in Fig.~\ref{fig:justification} (b). This property additionally helps the model reason about spatial interaction from image evidence.

\myparagraph{Sensor-Agnostic Planning in the Image Plane.}
Leveraging this theoretical foundation, the {Planning} phase fully exploits the native visual grounding capabilities of the VLM to perform high-level reasoning and 2D trajectory prediction. Let $\mathbf{I}=\{I_i\in \mathbb{R}^{H \times W \times 3}\}_{i=1}^{N_\text{view}}$ represent the input multi-view images and $\mathbf{C}$ denote the user command (including ego-status). We formulate the driving task as a conditional generation problem exclusively on the image plane $\mathcal{P}_\text{img}$. The VLM, denoted as $\Phi$, processes the visual and textual inputs to sequentially generate detected objects $\mathcal{\hat{B}}_\text{2D}$, reasoning with bounding boxes $\mathcal{\hat{R}}$, meta-action $\hat{A}$, and future waypoints $\mathcal{\hat{T}}_\text{2D}$ in pixel coordinates in an auto-regressive manner:
\begin{equation}
\label{eq:autoregressive_chain}
\begin{aligned}
   (\mathcal{\hat{B}}_\text{2D}, \mathcal{\hat{R}}, \hat{A}, \mathcal{\hat{T}}_\text{2D} \mid \mathbf{I}, \mathbf{C}) \sim & \ \prod_{t=1}^{T} p_\Phi\big((\hat{u}_t, \hat{v}_t) \mid \mathbf{I}, \mathbf{C}, \mathcal{\hat{B}}_\text{2D}, \mathcal{\hat{R}}, \hat{A}, \mathcal{\hat{T}}_{\text{2D}, <t}\big) \\
   & \cdot p_\Phi(\hat{A} \mid \mathbf{I}, \mathbf{C}, \mathcal{\hat{B}}_\text{2D}, \mathcal{\hat{R}}) \\
   & \cdot p_\Phi(\mathcal{\hat{R}} \mid \mathbf{I}, \mathbf{C}, \mathcal{\hat{B}}_\text{2D}) \\
   & \cdot p_\Phi(\mathcal{\hat{B}}_\text{2D} \mid \mathbf{I}, \mathbf{C})
\end{aligned}
\end{equation}
where $(\hat{u}_t, \hat{v}_t)$ represents the planned image-space waypoint on $\mathcal{P}_\text{img}$ at future time step $t$.
By formulating perception and trajectory prediction as 2D-to-2D prediction tasks, the planning phase is entirely \textit{sensor-agnostic}. We can merge heterogeneous calibrated datasets (e.g., nuScenes, Waymo, and internal proprietary data) into a single unified training corpus for scalable training without requiring the VLA to learn dataset-specific 2D-to-3D mappings. Deterministic lifting still requires the target camera parameters at inference, so weakly calibrated or uncalibrated 3D planning remains outside the scope of this work.

Notably, unlike previous VLAs that either treat perception only as a proxy task for 2D-to-3D feature projection~\cite{wang2025omnidrive,fu2025orion,opendrivevla,liu2025drivepi} or generate coarse textual descriptions (which are prone to ambiguous object referencing)~\cite{yuan2025autodrive,chi2025impromptu,autovla}, we introduce \textit{reasoning with bounding boxes}, where PixelPilot explicitly attends to specific Regions of Interest (RoI) by conditioning its semantic reasoning $\mathcal{\hat{R}}$ on the preceding perception outputs $\mathcal{\hat{B}}_\text{2D}$ (e.g., \texttt{<think> The car at [363, 239, 402, 274] is directly ahead and appears close, so I need to reduce \\speed </think>} in Fig.~\ref{fig:paradigm}). In this way, we establish an interpretable causal chain: the model must explicitly \textit{perceive} a visual anchor, \textit{reason} about its spatial relationship, and then \textit{act and plan}, which encourages predicted trajectories to be anchored to verifiable visual evidence rather than coarse language priors or trivial ego-status. 

\myparagraph{Deterministic Lifting to the 3D World.}
Once the 2D trajectory $\mathcal{\hat{T}}_\text{2D}$ is generated, the Lifting phase converts it into 3D trajectories. Since 3-second planning in autonomous driving typically occurs on a local plane~\cite{lim2021patchwork,lee2022patchwork++}, we empirically set the height of this plane as $Z = -h$.
Let $K$ be the camera parameters transforming world coordinates to image coordinates. For planned 2D waypoints $\mathcal{\hat{T}}_\text{2D} = \{ [\hat{u}_t, \hat{v}_t, 1] \}_{t=1}^{T}$ (in homogeneous coordinates), we seek their corresponding 3D points $\mathcal{\hat{T}}_{3D} = \{ [\hat{x}_t, \hat{y}_t, -h] \}_{t=1}^{T}$ in the ego-vehicle coordinate via the mapping function $\Psi$, which is formulated as a homography transformation or geometric ray-casting:
\begin{equation}
    \mathcal{\hat{T}}_{3D} = \Psi(\mathcal{\hat{T}}_{2D}, K, h)
\end{equation}
Specifically, we cast a ray from the camera center through the pixel $(u, v)$ and compute its intersection with the local plane where $z = -h$. 

Decoupling 2D-to-3D lifting from optimization ensures that absolute ego-speed cannot be trivially mapped to pixel displacement without contextual depth, facilitating the exploitation of visual cues rather than ego-status.
Notably, since the sensor-specific 2D-to-3D lifting occurs only during inference, the trained VLA can generalize to another calibrated vehicle by simply updating the camera parameters in the subsequent lifting phase. This formulation does not reject 3D priors; instead, such priors can be injected into the lifting stage while keeping the learnable policy in scalable visual space.

\begin{figure*}[!t]
    \centering
    \includegraphics[width=0.95\linewidth]{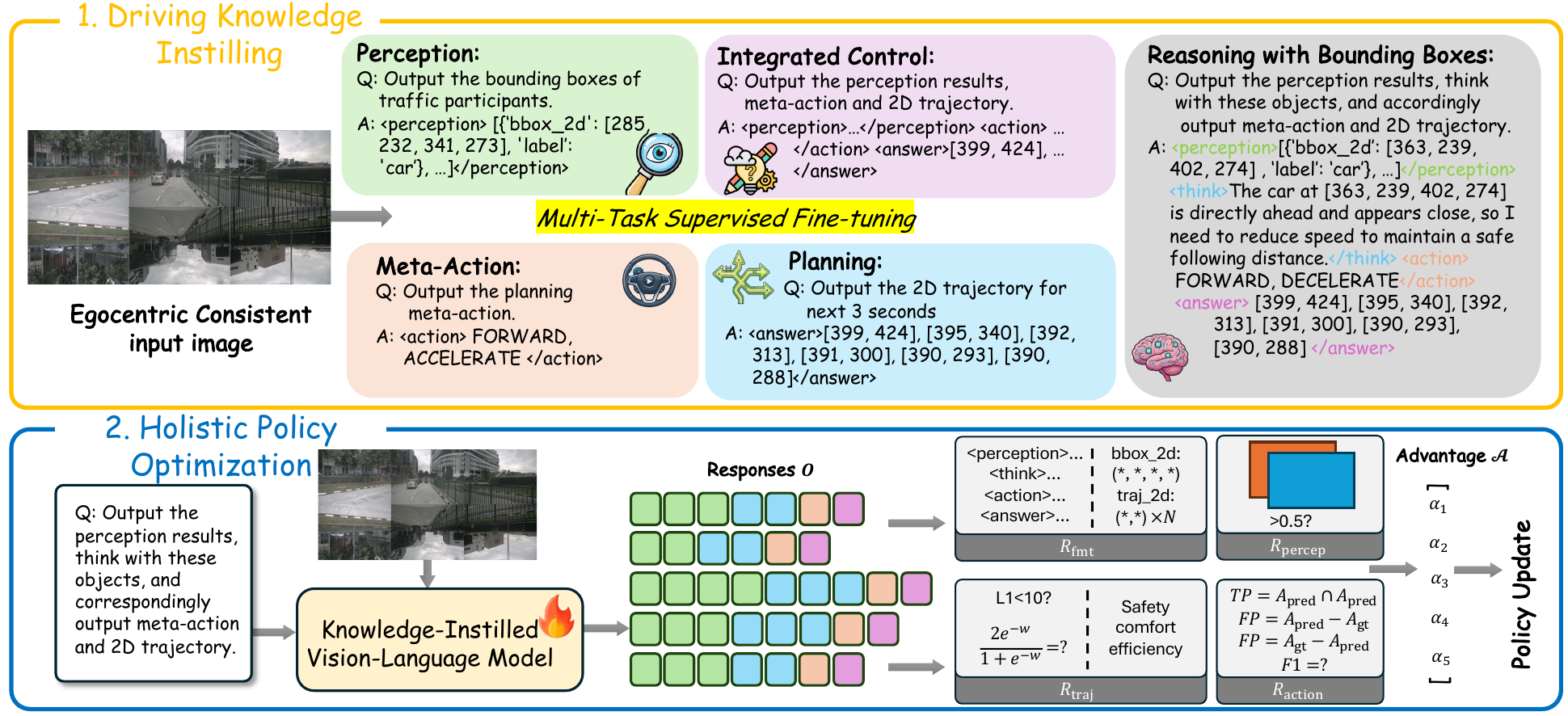}
    \caption{{{Training pipeline of PixelPilot.}} We propose a knowledge-instilled policy learning strategy. The first stage instills foundational driving knowledge into the VLA by multi-task SFT. The second stage assigns rewards on the holistic VLA pipeline for coherence.
    }
    \label{fig:training}
\end{figure*}

\subsection{Knowledge-Instilled Policy Learning}
\label{sec:optimize}
To effectively implement the decoupled planning and lifting paradigm, we first preprocess the input multi-view images for ego-centric consistency. Then, as illustrated in Fig.~\ref{fig:training}, we utilize a two-stage learning strategy that first instills driving knowledge via supervised fine-tuning (SFT) and subsequently aligns the generated Chain-of-Thought with final trajectory prediction by applying dense rewards from perception to planning via Reinforcement Learning (RL).

\begin{figure*}[!t]
    \centering
    \includegraphics[width=0.95\linewidth]{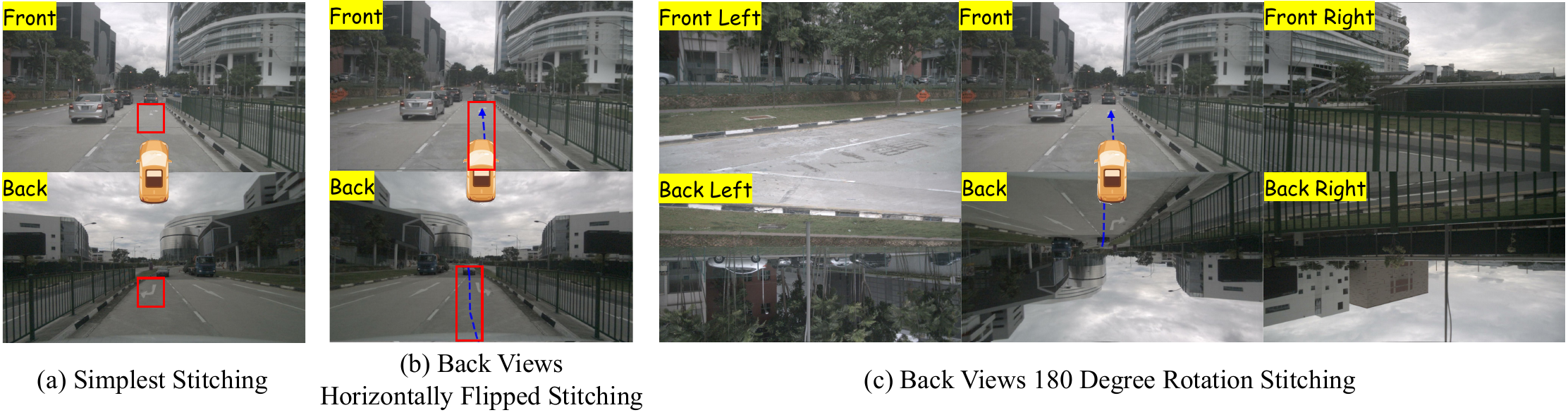}
    \caption{Different image preprocessing methods for VLAs.
    }
    \label{fig:stitch}
\end{figure*}

\myparagraph{Ego-Centric Consistency Preprocessing.} 
To enable coherent 2D driving across multiple cameras, we address a critical flaw in prevailing VLA multi-view compositions. Typically, front and back camera inputs are arranged into two separate rows, introducing severe \textit{egocentric inconsistency}, such as lane marking misalignment when using the simplest stitching (Fig.~\ref{fig:stitch} (a)). While using horizontally flipped stitching resolves the lane marking misalignment, it generates discontinuous trajectory planning (Fig.~\ref{fig:stitch} (b)). To rectify this problem and construct a continuous 2D planning surface, we rotate the back view by 180 degrees. This spatial correction preserves egocentric consistency, generating a visually continuous road plane and a cohesive trajectory across the entire multi-camera image collage (Fig.~\ref{fig:stitch} (c)).

\myparagraph{Stage 1: Driving Knowledge Instilling via SFT.} 
The initial stage utilizes Supervised Fine-Tuning (SFT) to instill foundational driving knowledge into the VLA. We train the model on a comprehensive multi-task dataset meticulously engineered to facilitate visual reasoning:

\noindent\textbf{\textit{(1) Perception.}} Previous VLAs~\cite{wang2025omnidrive,fu2025orion,opendrivevla,liu2025drivepi} use perception as a proxy task for 2D-to-3D feature projection, leading to limited data scaling and visual reasoning. On the contrary, we fine-tune the VLA to directly output the 2D bounding boxes in an autoregressive manner, providing spatial anchors for reasoning and planning.
To generate 2D bounding box labels $\mathcal{B}_\text{2D}$, we follow common practice~\cite{wang2023exploring, tang2024simpb} and project the ground-truth 3D bounding boxes onto the corresponding 2D camera image planes. 

\noindent\textbf{\textit{(2) Meta-Action Prediction.}} 
We cultivate an understanding of high-level semantic driving intentions using meta-actions $A$, which comprise a lateral component $A_\text{lat}$ (TURN LEFT, TURN RIGHT, FORWARD) and a longitudinal component $A_\text{lon}$ (ACCELERATE, DECELERATE, KEEP, STOP). This meta-action provides a prior for final trajectory prediction.

\noindent\textbf{\textit{(3) Planning.}} We use 2D trajectory data to predict precise motion planning. The 2D trajectory annotation $\mathcal{T}_\text{2D}$ is obtained using the same projection procedure as perception. Notably, although the perception and trajectory labels are constructed through 3D-to-2D projection to leverage the existing 3D-labeled datasets, our PixelPilot does not necessitate 3D bounding box or trajectory collection in real-world deployment.

\noindent\textbf{\textit{(4) Integrated Control.}} We employ three-step perception-action-planning sequences to foster multi-task learning and establish the causal link between observation and execution.

\noindent\textbf{\textit{(5) Reasoning with Bounding Boxes.}} 
We train our PixelPilot with {reasoning with bounding boxes} to fully exploit the visual cues in images. To generate the reasoning data, we first evaluate the baseline SFT model on the training set to identify failure cases. For these cases, we employ the Qwen-VL-Max model and provide it with a structured prompt containing the input image $\mathbf{I}$, ground-truth action $A$, and the 2D bounding boxes $\mathcal{B}_\text{2D}$ of surrounding agents to synthesize a causal reasoning chain that logically justifies the ground-truth action based on the perceived bounding boxes.
Then, we feed the original input combined with the generated reasoning back into the SFT model. A sample is strictly selected only if the inclusion of reasoning leads to a tangible improvement, i.e., correcting the predicted action and reducing the trajectory L2 error. 

Through the SFT stage, the model possesses a strong behavioral prior and can perform basic driving tasks. Please refer to the Supplementary Material for detailed 2D labels and reasoning data generation.

\myparagraph{Stage 2: Holistic Policy Optimization via RL.} 
Building on the SFT-initialized model, this phase employs the Group Relative Policy Optimization (GRPO)~\cite{grpo} algorithm to incentivize the VLA's higher-level reasoning and decision-making. 
Crucially, unlike previous RL practices~\cite{autovla,yuan2025autodrive} in autonomous driving that typically compute sparse rewards exclusively on the final planning output (e.g., trajectory deviation or collision penalties), our optimization strategy calculates dense, intermediate rewards on verifiable outputs across the auto-regressive pipeline, including perception, meta-action, and trajectory. We do not assign semantic rewards to free-form reasoning with bounding boxes, preserving its flexibility and reducing shortcut learning. By evaluating the response quality from initial perception down to final planning, this holistic reward structure yields two fundamental advantages. First, it mitigates the severe credit assignment problem inherent in long-horizon VLM generation by providing dense supervisory signals at each logical step. Second, it enforces the causal chain established during SFT. By simultaneously optimizing intermediate visual perception and final spatial planning, we encourage the final driving intent to remain anchored in accurate visual scene understanding rather than trivial ego-status.

\noindent\textbf{\textit{Format Reward}} $R_\text{fmt}$. It is designed to ensure a strict and hierarchical output structure. It consists of three parts: (1) {A structural reward} of 1.0 is granted if the generated responses are in the form: \texttt{"<perception>bboxes here</perception> <think>thinking with bboxes here</think> <action>meta-action here \\</action> <answer>trajectory here</answer>"}, otherwise 0. (2) {A perception format reward} ensures that each bounding box is complete. (3) {A trajectory format reward} of 1.0 is allocated for trajectory completeness and consistency.

\noindent \textbf{\textit{Perception Reward}} $R_\text{percep}$. We employ an Intersection over Union (IoU)-based reward. Specifically, we use the Hungarian algorithm to find the matched predicted and ground truth (GT) bounding boxes and compute the mean IoU between them as the final perception reward.

\noindent \textbf{Meta-Action Reward} $R_\text{action}$. We evaluate the accuracy of the predicted high-level meta-action using an F1-score-based reward, balancing precision and recall against the ground-truth action set. 

\noindent \textbf{Trajectory Reward} $R_\text{traj}$. 
To prioritize adherence to the 2D intent, we assign an L1 reward ($R_\text{traj-L1} = 1.0$) if the absolute pixel distance between the predicted and ground-truth 2D waypoints is below 10 pixels. We also apply a sigmoid-scaled L2 reward: $R_\text{traj-L2}=\frac{2e^{-w}}{1+e^{-w}}$, where $w$ is the L2 distance. Moreover, we employ the Predictive Driver Model Score (PDMS)~\cite{dauner2024navsim} $R_\text{PDMS}$ to assess the safety, comfortness, and efficiency of closed-loop predicted trajectories.

The final accumulated reward is the weighted sum of these terms, ensuring coherence among visual perception, logical intent, and spatial planning via dense rewards.
\begin{equation}
    R_{\text{acc}} = \lambda_{\text{fmt}} \cdot R_{\text{fmt}}
+ \lambda_{\text{percep}} \cdot R_{\text{percep}}
+ \lambda_{\text{action}} \cdot R_{\text{action}}
+ \lambda_{\text{traj}} \cdot R_{\text{traj}} \, .
\end{equation}

\section{Experiments}
\subsection{Implementation and Metrics}
\myparagraph{Datasets.} 
We evaluate the perception, action prediction, and open-loop performance on nuScenes~\cite{caesar2020nuscenes}. 
For closed-loop evaluation, we use the Bench2Drive benchmark~\cite{jia2024bench2drive} in the CARLA simulator.
Besides, we incorporate the Waymo Open dataset~\cite{sun2020scalability} to demonstrate the data scalability of our PixelPilot.

\myparagraph{Implementation Details.}
We use the powerful open-source VLM Qwen2.5-VL~\cite{Qwen2.5VL} as the base model for PixelPilot. 
For SFT, i.e., stage 1, we fine-tune the model for 2 epochs in a multi-task manner to instill the model with knowledge about perception, reasoning, high-level meta-action making, and trajectory prediction. During Stage 2, reinforcement learning, we fine-tune the model trained by SFT using GRPO~\cite{grpo} to incentivize the reasoning ability of the trained policy. The number of completions in GRPO is set to 8. All the experiments were conducted on eight GPUs with 80 GB memory each.
For the main open-loop experiments, PixelPilot is trained on nuScenes (28.1k samples) and Waymo (23.8k samples) with SFT and RL. For Bench2Drive closed-loop evaluation, following Orion~\cite{fu2025orion}, we train on Bench2Drive (274.5k samples) with SFT only, plan a 3-second trajectory at 2 Hz from current multi-view images, and convert the predicted trajectory to vehicle control with a PID controller. During decoding, we set temperature to 1.0, top-$p$ to 0.5, and top-$k$ to 20.

\myparagraph{Evaluation Metrics.} For perception, we report the mAP for 2D object detection. For meta-action, we use the F1-score for all lateral and longitudinal meta-action classes. For open-loop planning, we employ the L2 error (in meters) between the predicted and ground-truth trajectories. Additionally, following \cite{bev-planner}, the collision rate and intersection rate are adopted to evaluate the safety of the trajectory. Since open-loop safety metrics can be affected by non-reactive followers and off-road intersections, we jointly report L2, collision, and intersection and further validate safety in the closed-loop benchmark.
For closed-loop performance, we report the success
rate, driving score, efficiency, and comfortness.

\begin{table}[t]
    \scriptsize
    \scalebox{0.93}{
        \begin{minipage}[b]{0.29\linewidth}
        \centering
        \makeatletter\def\@captype{table}
        \setlength{\tabcolsep}{1mm}{
            \caption{2D Object detection on nuScenes.}
            \label{tab:percep}
            \begin{tabular}{l|c}
            \toprule
            Method  & mAP $\uparrow$ \\
            \midrule 
            StreamPETR~\cite{wang2023exploring} & 46.5 \\
            MV2D~\cite{mv2d} & 52.3  \\
            DeformableDETR~\cite{deformabledetr} & 50.2 \\
            SimPB~\cite{tang2024simpb} & {54.1} \\
            \midrule 
            \cellcolor[gray]{.9}{PixelPilot} & \textbf{54.2}  \\
            \bottomrule
            \end{tabular}}
        \end{minipage}

        \quad
        \hspace{8pt}
        \begin{minipage}[b]{0.69\linewidth}
        \centering
        \makeatletter\def\@captype{table}
        \setlength{\tabcolsep}{1mm}{
            \caption{High-level meta-action prediction on nuScenes. ${\text{\textdagger}}$ indicates trained on nuScenes.}
            \label{table:action}
            \begin{tabular}{l|ccc|cccc}
            \toprule
            \multirow{2}{*}{Method} & \multicolumn{3}{c|}{Lateral $($F1$)$ $\uparrow$} & \multicolumn{4}{c}{Longitudinal $($F1$)$ $\uparrow$} \\
            & forward & left & right & keep & acc. & dec. & stop \\
            \midrule
            Qwen2.5VL-7B  & 64.67 & 24.15 & 30.85 & 40.73 & 55.14 & 51.41 & 41.82  \\
            \midrule
            Qwen2.5VL-7B${\text{\textdagger}}$ & {94.46} & {63.00} & {67.01} & {57.62} & {74.35} & {77.10} & {75.00} \\
            \midrule
            \cellcolor[gray]{.9}PixelPilot & \textbf{96.82} & \textbf{75.51} & \textbf{75.71} & \textbf{61.23} & \textbf{81.76} & \textbf{80.19} & \textbf{81.80} \\
            \bottomrule
            \end{tabular}}
        \end{minipage}}
\end{table}

\subsection{Main Results}
\noindent \textbf{Perception.} We evaluate the perception capability of PixelPilot in Tab.~\ref{tab:percep}. 
Notably, our VLA model achieves state-of-the-art performance, even compared to specialized 2D object detectors trained on nuScenes~\cite{wang2023exploring, mv2d,deformabledetr, tang2024simpb}, despite not being optimized only for 2D object detection.

\myparagraph{Meta-Action.} The performance of PixelPilot on meta-action prediction is detailed in Table~\ref{table:action}. Compared to the base Qwen2.5VL-7B, PixelPilot demonstrates superior performance across all lateral (Path) and longitudinal (Speed) action categories when evaluated by F1-score, underscoring the effectiveness of our proposed training methodology.

\begin{table*}[t]
\setlength{\tabcolsep}{0.004\linewidth}
\centering
\caption{{Open-loop trajectory planning of VLAs on nuScenes. As FSDrive~\cite{zeng2025futuresightdrive}} uses private data for training and AutoDrive-R$^2$~\cite{yuan2025autodrive} explicitly calculates the 3D trajectory from ego status via kinematics, we mark them as \textcolor{gray}{gray}.}
\resizebox{0.9\linewidth}{!}{
    \begin{tabular}{l|cccc|cccc|cccc}
    \toprule
    \multirow{2}{*}{Method} &
    \multicolumn{4}{c|}{L2 (m) $\downarrow$} & 
    \multicolumn{4}{c|}{Collision (\%) $\downarrow$} &
    \multicolumn{4}{c}{Intersection (\%) $\downarrow$} \\
     & 1s & 2s & 3s &\cellcolor{gray!30}Avg. & 1s & 2s & 3s& 
    \cellcolor{gray!30}Avg. & 1s & 2s & 3s &\cellcolor{gray!30}Avg.\\
    \midrule
    DriveVLM~\cite{drivevlm} & 0.18 & 0.34 & 0.68 & 0.40 & 0.10 & 0.22 & 0.45 & 0.27 & - & - & - & -\\
    EMMA~\cite{hwang2024emma} & 0.14 & 0.29 & 0.54 & 0.32 & - & - & - & - & - & - & - & - \\
    OmniDrive~\cite{wang2025omnidrive} & {0.14} & {0.29} & {0.55} & {0.33} & \textbf{0.00} & {0.13} & 0.78 & {0.30} & 0.56 & 2.48 & 5.96 & 3.00 \\
    OpenDriveVLA~\cite{opendrivevla} & 0.15 & 0.31 & 0.55 & 0.33 & 0.01 & \textbf{0.08} & \textbf{0.21} & \textbf{0.10} & - & - & - & - \\
    Imprompt-VLA~\cite{chi2025impromptu} & \textbf{0.13} & 0.27 & 0.53 & \textbf{0.30} & - & - & - & - & - & - & - & - \\
    \textcolor{gray}{FSDrive~\cite{zeng2025futuresightdrive}} & \textcolor{gray}{0.14}  & \textcolor{gray}{0.25} & \textcolor{gray}{0.46} & \textcolor{gray}{0.28} & \textcolor{gray}{0.07} & \textcolor{gray}{0.12} & \textcolor{gray}{1.02} & \textcolor{gray}{0.40} & \textcolor{gray}{-} & \textcolor{gray}{-} & \textcolor{gray}{-} & \textcolor{gray}{-} \\
    AutoVLA~\cite{autovla} & 0.21 & 0.38 & 0.60 & 0.40 & 0.13 & 0.18 & 0.28 & 0.20 & - & - & - & - \\
    \textcolor{gray}{AutoDrive-R$^2$~\cite{yuan2025autodrive}} & \textcolor{gray}{0.11} & \textcolor{gray}{0.19} & \textcolor{gray}{0.29} & \textcolor{gray}{0.20} & \textcolor{gray}{-} & \textcolor{gray}{-} & \textcolor{gray}{-} & \textcolor{gray}{-} & \textcolor{gray}{-} & \textcolor{gray}{-} & \textcolor{gray}{-} & \textcolor{gray}{-} \\
    DrivePI~\cite{liu2025drivepi} & 0.19 & 0.36 & 0.64 & 0.40 & 0.00 & 0.05 & 0.28 & 0.11 & - & - & - & - \\
    \midrule
    \cellcolor[gray]{.9}PixelPilot & \textbf{0.13} & \textbf{0.26} & \textbf{0.51} & \textbf{0.30} &  \textbf{0.00} & {0.11} & {0.65} & {0.25} & \textbf{0.51} & \textbf{1.53} & \textbf{3.26} & \textbf{1.77} \\
    
    \bottomrule
    \end{tabular}}
    \label{table:open}
\end{table*}

\myparagraph{Open-Loop Planning.}
In the open-loop trajectory prediction, as detailed in Table~\ref{table:open}, PixelPilot achieves a state-of-the-art average L2 error of 0.30 m.  
Compared with AutoVLA~\cite{autovla}, PixelPilot reduces the average L2 error from 0.40 m to 0.30 m, corresponding to a 25\% relative reduction.
The safety and feasibility of the planned trajectories are evaluated by the collision rate, where PixelPilot consistently achieves competitive performance.
For Intersection, which measures trajectory feasibility with respect to the drivable area, PixelPilot again shows strong results. It obtains the best average rate (1.77\%) among its VLA-based peers. This comprehensive performance in safety-critical metrics validates the effectiveness of our PixelPilot.

\myparagraph{Closed-Loop Planning.}
We conduct the closed-loop test on Bench2Drive~\cite{jia2024bench2drive} in CARLA simulator at 2 Hz. As shown in Tab.~\ref{tab:close}, PixelPilot achieves the best overall driving score and success rate, validating our decoupled planning and lifting paradigm. 
Among VLA-based methods, PixelPilot also achieves the best efficiency and the second-best comfortness. Compared with the traditional planner VAD~\cite{vad}, PixelPilot remains competitive in efficiency and comfortness while substantially improving the core driving safety metrics by 36.79 driving score and 43.87 success rate.

\begin{table*}[t]
\setlength{\tabcolsep}{0.004\linewidth}
\centering
\caption{Closed-loop planning on Bench2Drive (CARLA).}
\resizebox{0.8\linewidth}{!}{
    \begin{tabular}{l|cccc}
    \toprule
    Method & Driving Score $\uparrow$ & Success Rate (\%) $\uparrow$ & Efficiency $\uparrow$ & Comfortness $\uparrow$ \\
    \midrule
    AD-MLP \cite{ego-mlp}            & 18.05           & 0.00                 &  48.45        & 22.63   \\
    UniAD-Base \cite{uniad}            & 45.81           & 16.36                & 129.21        & 43.58   \\
    VAD \cite{vad}                     & 42.35           & 15.00                & \textbf{157.94}        & \textbf{46.01}    \\
    TCP-traj \cite{wu2022trajectory}            & 59.90           & 30.00                & 76.54         & 18.08 \\
    DriveAdapter \cite{jia2023driveadapter}     & 64.22           & 33.08                & 70.22         & 16.01  \\
    Orion~\cite{fu2025orion} & 77.74 & 54.62 & 151.48 & 17.38  \\ 
    AutoVLA~\cite{autovla} & {78.84} & {57.73} & {146.93} & {39.33} \\
    \midrule
    \cellcolor[gray]{.9}PixelPilot & \textbf{79.14} & \textbf{58.87} & 153.26 & 38.01  \\
        \bottomrule
    \end{tabular}}
    \label{tab:close}
\end{table*}

\begin{figure*}[!t]
    \centering
    \includegraphics[width=0.95\linewidth]{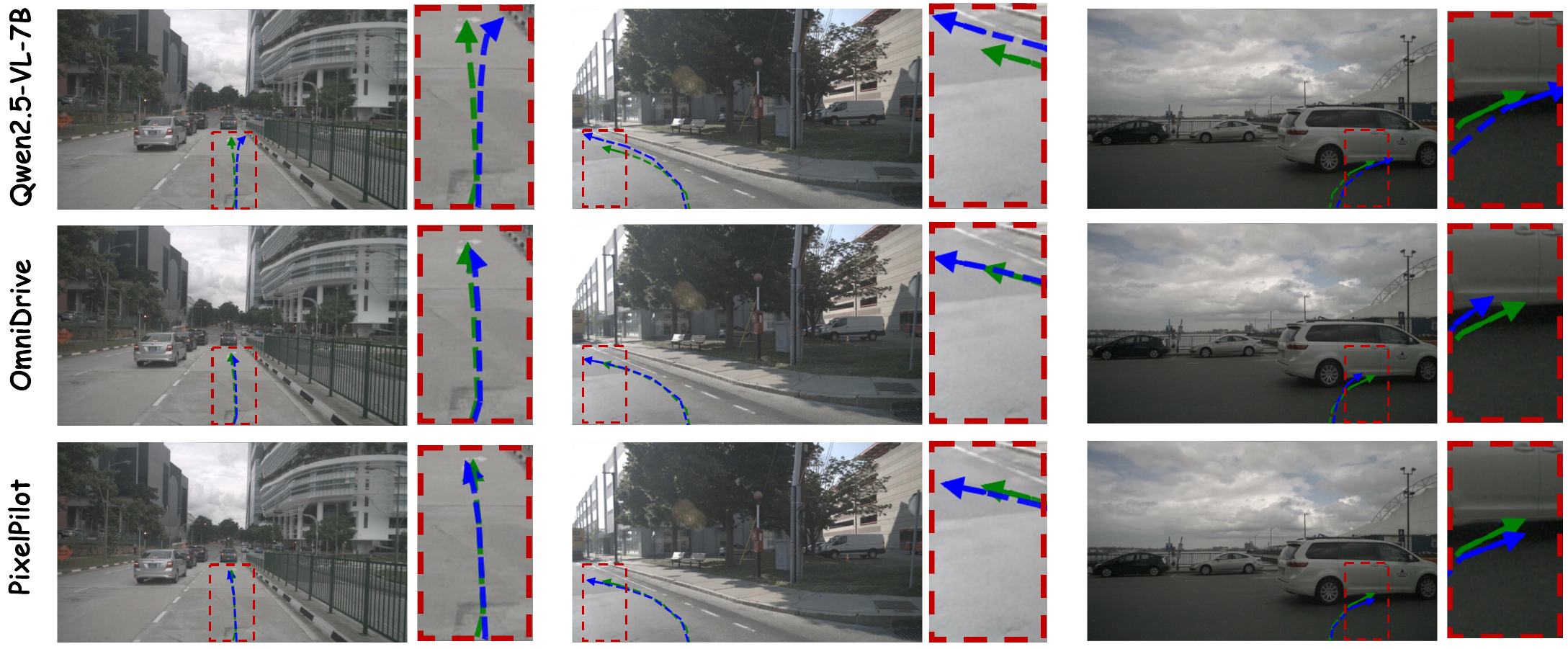}
    \caption{{Visualization of 2D trajectories across Qwen2.5-VL-7B, OmniDrive, and our PixelPilot on the nuScenes validation dataset.} The predicted and ground-truth trajectories are depicted in blue and green, respectively.
    }
    \label{fig:vis}
\end{figure*}
\subsection{Visualization}
Fig.~\ref{fig:vis} presents a qualitative comparison of our method against other approaches on the nuScenes dataset. Notably, Qwen2.5-VL-7B struggles to generate accurate predictions, exhibiting significant trajectory deviations, particularly in turning scenarios. Although OmniDrive delivers satisfactory performance, its trajectories are often overly aggressive, with predicted speeds substantially exceeding ground truth speeds. In contrast, our method, PixelPilot, consistently generates reliable plans that closely align with the ground truth.

\subsection{Ablation Study}
\noindent \textbf{Two Stage Training Pipeline.} To validate our two-stage knowledge-instilled learning strategy, we conducted an ablation study (Table~\ref{tab:two_stage}) by training variants with only Supervised Fine-Tuning (SFT) or Reinforcement Learning (RL). We find that while both methods provide performance gains, the SFT-only variant surpasses the RL-only model. This suggests that RL, on its own, is inefficient at navigating the vast search space of our task's structured multi-step reasoning process (perception → reasoning → meta-action → planning). SFT is therefore essential for instilling the model with a coherent policy and a foundational understanding of the causal chain required. The superior performance of the complete PixelPilot model, which combines both stages, confirms that our hybrid approach is critical: SFT provides the necessary knowledge foundation, which RL then effectively refines to achieve coherent optimal results.

\begin{table}[t]
    \scriptsize
    \scalebox{1}{
        \begin{minipage}[b]{0.33\linewidth}
        \centering
        \makeatletter\def\@captype{table}
        \setlength{\tabcolsep}{1mm}{
            \caption{Ablation on learning strategy.}
            \label{tab:two_stage}
            \begin{tabular}{lc}
        \toprule 
        {Method} & Avg. L2 $\downarrow$ \\
        \midrule
        Qwen2.5-VL-7B        & 1.92\\
        Qwen2.5-VL-7B + SFT  & 0.46 \\
        Qwen2.5-VL-7B + RL   & 0.49 \\
        \bottomrule
        \end{tabular}}
        \end{minipage}

        \quad
        \hspace{8pt}
        \begin{minipage}[b]{0.29\linewidth}
        \centering
        \makeatletter\def\@captype{table}
        \setlength{\tabcolsep}{1mm}{
            \caption{Ablation on preprocessing and SFT.}
            \label{tab:sft}
            \begin{tabular}{lc}
        \toprule 
        {Method} & Avg. L2 $\downarrow$ \\
        \midrule
        Front View & 0.40 \\
        w/o. Ego. Cons. & 0.43 \\
        w/o. Multi-Task & 0.45 \\
        \bottomrule
        \end{tabular}}
        \end{minipage}
        
        \quad
        \hspace{8pt}
        \begin{minipage}[b]{0.25\linewidth}
        \centering
        \makeatletter\def\@captype{table}
        \setlength{\tabcolsep}{1mm}{
            \caption{Ablation on RL.}
            \label{tab:rl}
            \begin{tabular}{lc}
        \toprule 
        {Method} & Avg. L2 $\downarrow$ \\
        \midrule
        w/o. $R_{\text{traj}}$   & 0.44 \\
        w/o. $R_{\text{percep}}$ & 0.41 \\
        w/o. $R_{\text{action}}$ & 0.40 \\
        \bottomrule
        \end{tabular}}
        \end{minipage}
        
        }
\end{table}

\noindent\textbf{Preprocessing and SFT.}
To ensure the model's correct understanding and reasoning, the training images are concatenated in an ego-centrically consistent way. Then, we SFT Qwen2.5-VL-7B on a multi-task dataset, including 2D object detection, reasoning with bounding boxes, meta-action prediction, and trajectory prediction. 
Table~\ref{tab:sft} shows that removing the ego-centric consistent input (`w/o. Ego. Cons.', i.e., simplest stitching) leads to a higher L2 error compared to only using the front view, demonstrating the effectiveness of preprocessing. Replacing the multi-task training with simple trajectory alignment (`w/o. Multi-Task') increases the average L2 error to 0.45 m, validating the superiority of the perception-to-planning CoT learning.

\noindent\textbf{Reinforcement Learning.}
We dissect the contributions of each component within our composite reward function used during the Reinforcement Learning (RL) stage. As shown in Table~\ref{tab:rl}, individually ablating trajectory reward $R_{\text{traj}}$, IoU-based perception reward $R_{\text{percep}}$, and action reward $R_{\text{action}}$ each leads to a discernible increase in the average L2 error. This confirms that these verifiable rewards all positively contribute to coherent CoT and final planning accuracy while avoiding over-constraint on free-form reasoning with bounding boxes. 

\noindent\textbf{Scalability and Generalization.}
When excluding the Waymo dataset from training, we observe a 0.06 increase in average L2 error, demonstrating the scalability brought by our decoupled planning and lifting paradigm. To empirically validate the generalization of PixelPilot across different camera configurations, we extended our trajectory prediction evaluation to the Waymo Open Dataset~\cite{sun2020scalability}, as reported in Tab.~\ref{tab:waymo}. Remarkably, despite the significant domain shift between the training source (nuScenes) and the target domain (Waymo), our method achieves competitive performance with an average L2 error of $0.4\text{m}$ in a zero-shot setting, i.e., without any dataset-specific fine-tuning. Additional controlled ablations on planning targets and 3D-prior injection are provided in the Supplementary Material.

\begin{table}[t]
    \scriptsize
    \scalebox{0.93}{
        \begin{minipage}[b]{0.45\linewidth}
        \centering
        \makeatletter\def\@captype{table}
        \setlength{\tabcolsep}{1mm}{
            \caption{Trajectory prediction on Waymo.}
            \label{tab:waymo}
            \begin{tabular}{lccccc}
            \toprule 
            \multirow{2}{*}{{Method}} & \multicolumn{4}{c}{{L2 Error $\downarrow$}} \\
            \cmidrule(lr){2-5}
             & 1s & 2s & 3s & Avg. \\
            \midrule
            Qwen2.5-VL-7B~\cite{Qwen2.5VL} & 1.66 & 1.82 & 2.92 & 2.13 \\
            DriveVLM~\cite{drivevlm} & 0.17 & 0.34 & 0.75 & 0.42 \\
            EMMA~\cite{hwang2024emma} & 0.12 & 0.28 & 0.61 & 0.34 \\
            EMMA+~\cite{hwang2024emma} & 0.11 & 0.25 & 0.53 & 0.30 \\
            \midrule
            \cellcolor[gray]{.9}{PixelPilot} & 0.16 & 0.32 & 0.71 & 0.40 \\
            \bottomrule
            \end{tabular}}
        \end{minipage}

        \quad
        \hspace{8pt}
        \begin{minipage}[b]{0.5\linewidth}
        \centering
        \makeatletter\def\@captype{table}
        \setlength{\tabcolsep}{1mm}{
            \caption{Ablation study on ego-status and input images.}
            \label{tab:ego_input}
            \begin{tabular}{lccc} 
            \toprule
            {{Method}} & {{Ego-Status}} & {Images} & Avg. L2$\downarrow$ \\
            \midrule
            \multirow{2}{*}{OmniDrive~\cite{wang2025omnidrive}} 
            & \ding{55} & \ding{51} & 1.98  \\ 
            & \ding{51} & \ding{55} & 0.36  \\ 
            \midrule
            \multirow{2}{*}{Imprompt-VLA~\cite{chi2025impromptu}}
            & \ding{55} & \ding{51} & 2.39  \\ 
            & \ding{51} & \ding{55} & 0.36  \\ 
            \midrule
            \multirow{2}{*}{PixelPilot} 
            & \ding{55} & \ding{51} & 0.71  \\ 
            & \ding{51} & \ding{55} & 0.90  \\ 
            \bottomrule
        \end{tabular}}
        \end{minipage}}
\end{table}

\noindent\textbf{Ego-Status vs Input Images.}
Table~\ref{tab:ego_input} investigates whether our PixelPilot relies on visual cues or trivial ego-status. Under single-modality ablations, OmniDrive~\cite{wang2025omnidrive} and Imprompt-VLA~\cite{chi2025impromptu} obtain substantially lower errors with ego-status only than with images only, whereas PixelPilot exhibits the opposite trend, revealing that the direct 2D-to-3D prediction paradigm predominantly relies on ego-status regardless of the input images. By decoupling the planning and lifting phases, PixelPilot achieves a lower L2 error of 0.71 when utilizing only visual inputs (ego-status masked), while yielding a higher error of 0.90 when using only ego-status. This validates that our PixelPilot successfully prevents the network from merely extrapolating trajectories via trivial kinematics and grounds its planning in robust visual understanding.

\begin{table}[t]
    \scriptsize
    \scalebox{0.93}{
        \begin{minipage}[b]{0.43\linewidth}
        \centering
        \makeatletter\def\@captype{table}
        \setlength{\tabcolsep}{1mm}{
            \caption{Ablation on local plane.}
            \label{tab:local}
            \begin{tabular}{lcccc} 
            \toprule
            \multirow{2}{*}{{Method}} & \multicolumn{3}{c}{{Avg. L2 $\downarrow$}} \\
            \cmidrule(lr){2-4}
            & 0.5m & 1m & 2m \\ 
            \midrule
            Imprompt-VLA~\cite{chi2025impromptu} & 0.30 & 0.32 & 0.33  \\
            \cellcolor[gray]{.9}PixelPilot & 0.30 & 0.31 & 0.34 \\
            \bottomrule
            \end{tabular}}
        \end{minipage}

        \quad
        \hspace{8pt}
        \begin{minipage}[b]{0.55\linewidth}
        \centering
        \makeatletter\def\@captype{table}
        \setlength{\tabcolsep}{1mm}{
            \caption{Trade-off between CoT and latency.}
            \label{tab:trade_off}
            \begin{tabular}{lccc} 
            \toprule
            {{Method}} & {{Avg. L2 $\downarrow$}} & Latency (s) \\
            \midrule
            w/o CoT & 0.41 & 0.23  \\
            Coarse CoT & 0.35 & 0.31  \\
            \cellcolor[gray]{.9}PixelPilot & 0.30 & 0.33 \\
            \bottomrule
            \end{tabular}}
        \end{minipage}}
\end{table}

\noindent\textbf{Local Plane Assumption.} Tab.~\ref{tab:local} shows that our PixelPilot achieves similar performance to the 2D-to-3D VLA~\cite{chi2025impromptu} on samples with height changes above 0.5, 1, and 2 meters in future 3s. Together with the small average 3-second road-height variation in nuScenes, this supports the local-plane approximation for short-horizon ego-centric planning, while multi-level roads and severe occlusions remain challenging cases.

\noindent\textbf{Trade-off between CoT and Latency.} Table~\ref{tab:trade_off} shows that PixelPilot, featuring a CoT from perception to planning, reduces Avg. L2 from 0.35 m to 0.30 m over coarse CoT with only a marginal latency increase from 0.31 s to 0.33 s. This corresponds to about 3 Hz inference, satisfying the 2 Hz Bench2Drive control frequency, and the deterministic lifting stage adds negligible overhead.

\section{Conclusion}
In this paper, we introduced PixelPilot, a vision-language-action model for end-to-end autonomous driving that decouples sensor-agnostic 2D planning from sensor-specific 3D lifting. By formulating perception, visual reasoning, and trajectory prediction as image-space tasks, PixelPilot enables scalable learning from heterogeneous calibrated driving datasets while reducing the tendency of direct 3D optimization to rely on trivial ego-status cues. To realize this paradigm, we propose a two-stage knowledge-instilled policy learning strategy that grounds reasoning in explicit 2D bounding boxes and uses dense rewards on verifiable intermediate outputs to encourage a coherent causal chain from perception to planning. Extensive experiments show that PixelPilot achieves strong open-loop and closed-loop performance, supporting 2D-first planning with deterministic lifting as a scalable and interpretable formulation for driving VLAs.
This formulation also clarifies the current scope of the approach. While deterministic lifting avoids learning dataset-specific 2D-to-3D mappings, it still relies on reliable camera geometry and ego-motion estimates at inference time. Moreover, our geometric formulation is tailored to short-horizon ego-centric driving under a local-plane approximation; strongly non-planar multi-level roads, severe occlusions, and weakly calibrated sensor setups remain challenging. Extending the framework to handle these conditions more robustly, while preserving its sensor-agnostic learning advantages, is an important direction for future work.

\section*{Acknowledgements}
This work was supported by the National Natural Science Foundation of China (Grant Nos. 62322113, 62376156), as well as the Shanghai Municipal Special Program for Basic Research on General AI Foundation Models (Grant No. 2025SHZDZX025G15).

\bibliographystyle{splncs04}
\bibliography{main}
\end{document}